\documentclass[conference]{IEEEtran}
\IEEEoverridecommandlockouts

\usepackage{siunitx}
\usepackage{multirow}
\usepackage{amsmath}
\usepackage{booktabs} % Include in preamble
\usepackage{graphicx}
\usepackage{csquotes}
\usepackage{subcaption}
\usepackage{algorithm2e}
\usepackage{paralist}
\usepackage{comment}
\usepackage[table,xcdraw]{xcolor}

\usepackage[style=ieee]{biblatex}
\usepackage[
    colorlinks=true,
    linkcolor=red,      
    citecolor=green,       
    urlcolor=blue,       
    bookmarks=true,       
    bookmarksopen=true,
    pdfstartview=FitH
]{hyperref}
\usepackage[section,numberedsection=autolabel]{glossaries}
\makeglossaries

\newglossaryentry{function}{name=function,%
    description={A single, self-contained piece of software, that performs a certain function}}

\newglossaryentry{feature}{name=feature,
    description={Composed of one or more functions connected together using a certain runtime environment, usually corresponds to a certain use-case}}

\newglossaryentry{runtime environment}{name=runtime environment,
    description={Communication middleware and virtualization mechanisms}}

\newglossaryentry{ECU}{name=ECU,
    description={Electronic Control Unit. It is an electronic device in a vehicle that is responsible for a single function}}

\newglossaryentry{TDD}{name=TDD,
    description={Test-Driven Development is a software development methodology that centers on the iterative creation of unit tests prior to the implementation of functional code~\cite{ref40:Beck2002} This approach mandates that a test case specifying the desired behavior of a code unit be written before the production code itself. As development progresses, the test suite continuously executes. New code is only written if it fulfills the requirements outlined in a failing test}}

\newglossaryentry{FDD}{name=FDD,
    description={Feature-Driven Development. It is a paradigm where the software system is iteratively developed in a series of steps, starting with an abstract model of the system, followed by extraction of a set of desired features, and ending with feature implementation and integration~\cite{ref41:Palmer2001}}}

\newglossaryentry{MBSE}{name=MBSE,
    description={Model-Based Systems Engineering is a formalized methodology within systems engineering that emphasizes using models as the primary means of information exchange and system representation~\cite{ref42:Incose2023}. This contrasts with traditional document-centric approaches. MBSE centers on creating and leveraging domain-specific models or metamodels, which capture system requirements, design, analysis, and verification elements throughout the development lifecycle}}

\newglossaryentry{contract}{name=contract,
    description={Design by contract is a software development methodology that emphasizes the explicit definition of formal contracts between software components~\cite{ref43:Mitchell2002}. These contracts specify preconditions (what must be true before a component is used), postconditions (what must be true after execution), and invariants (conditions that must always hold true). Design by contract can be enforced through runtime assertions, unit tests, or even integrated into a programming language's syntax. This approach enhances software reliability, eases debugging, and facilitates code comprehension}}

\newglossaryentry{metamodel}{name=Metamodel,
    description={Defines the language of system description by specifying abstract entities that are part of the system, a set of possible relations between them, and their attributes}}

\newglossaryentry{instance model}{name=Instance model,
    description={A model generated from the given metamodel, populated with actual objects with concrete attribute values; an implementation of the system described in the language of the metamodel}}

\newglossaryentry{OMG}{name=OMG,
    description={Object Management Group}}

\newglossaryentry{LLM}{name=LLM,
    description={Large Language Model}}

\newglossaryentry{OCL}{name=OCL,
    description={Object Constraint Language}}

\newglossaryentry{RACE}{name=RACE,
    description={Centralized Platform Computer Based Architecture for Automotive Applications}}

\newglossaryentry{Ecore}{name=Ecore,
    description={Language of the metamodel used in Eclipse Modeling Framework}}

\newglossaryentry{OEM}{name=OEM,
    description={Original Equipment Manufacturer}}

\makeatletter % changes the catcode of @ to 11
\newcommand{\linebreakand}{%
  \end{@IEEEauthorhalign}
  \hfill\mbox{}\par
  \mbox{}\hfill\begin{@IEEEauthorhalign}
}
\makeatother % changes the catcode of @ back to 12

\begin{document}

%\title{Bridging Perception, Reasoning and Action: LLM-Augmented Multi-Sensor Fusion for Adaptive Autonomous Driving\\\vspace*{20pt} \normalsize  \today{}}
\title{A Unified Perception-Language-Action Framework for Adaptive Autonomous Driving
%\\
}

% Emphasizing Real-Time Efficiency:
%  "Real-Time Autonomous Driving via Multi-Sensor Fusion: A Lightweight Framework Combining Vision-Language-Action Models and LLMs"

% Highlighting Sensor-Model Synergy:
% "FusionDrive: Enhanced Decision-Making for Autonomous Systems Through Multi-Sensor Data Fusion and Vision-Language-Action-Language Model Integration"

% Focusing on Hybrid Intelligence:
%"Bridging Perception, Language, and Action: A Compact Real-Time Driving Agent Leveraging Multi-Sensor Fusion and LLM-Augmented VLA Architectures"
%yi zhang, orcid: 0009-0005-3678-0795
\author{
\IEEEauthorblockN{Yi Zhang, Erik Leo Haß, Kuo-Yi Chao, Nenad Petrovic, Yinglei Song, Chengdong Wu and Alois Knoll}
\IEEEauthorblockA{\textit{Chair of Robotics, Artificial Intelligence and Embedded Systems} \\
\textit{Technical University of Munich (TUM)}\\
Munich, Germany \\
\{yi1228.zhang, erik-leo.hass, kuoyi.chao, nenad.petrovic, yinglei.song, chengdong.wu, k\}@tum.de}
}
%\date{\today}

\maketitle

\begin{abstract}
Autonomous driving systems face significant challenges in achieving human-like adaptability, robustness, and interpretability in complex, open-world environments. These challenges stem from fragmented architectures, limited generalization to novel scenarios, and insufficient semantic extraction from perception. To address these limitations, we propose a unified Perception-Language-Action (PLA) framework that integrates multi-sensor fusion (cameras, LiDAR, radar) with a large language model (LLM)-augmented Vision-Language-Action (VLA) architecture, specifically a GPT-4.1-powered reasoning core. This framework unifies low-level sensory processing with high-level contextual reasoning, tightly coupling perception with natural language-based semantic understanding and decision-making to enable context-aware, explainable, and safety-bounded autonomous driving. Evaluations on an urban intersection scenario with a construction zone demonstrate superior performance in trajectory tracking, speed prediction, and adaptive planning. The results highlight the potential of language-augmented cognitive frameworks for advancing the safety, interpretability, and scalability of autonomous driving systems.

\end{abstract}

\section{Introduction}
\label{sec:intro}
%, which, if effectively integrated, could revolutionize autonomous driving systems by enabling dynamic, context-aware decision-making in real-world uncertainties.
%One promising direction lies in multi-sensor fusion, where LiDAR, cameras, and radar collectively enhance perception robustness. However, while these technologies improve raw data acquisition, the critical gap remains: bridging perceptual inputs with contextual reasoning and actionable outputs. Vision-Language models attempt to address this by grounding visual data in textual commands, yet they often lack the situational awareness and adaptability needed for complex driving scenarios. Meanwhile, Large Language Models (LLMs) have demonstrated exceptional capabilities in reasoning, planning, and interactive understanding—qualities that, if effectively integrated, could revolutionize decision-making in autonomous systems by enhancing their ability to interpret and respond to dynamic, real-world conditions.
Autonomous driving systems are rapidly evolving, yet fundamental challenges persist in achieving human-like cognitive fidelity during the integration of perception, decision-making, and control under stochastic real-world conditions. Unlike humans, current systems struggle to seamlessly integrate sensory inputs with contextual cues, hindering their ability to manage uncertainties in complex, dynamic environments. This limitation restricts adaptability in partially observable settings, such as urban traffic, where comprehensive scene understanding is critical. These challenges highlight the need for integrated frameworks that emulate human cognitive adaptability, enabling robust, context-aware decision-making.

A promising approach to address these challenges is multi-sensor fusion, integrating LiDAR, cameras, and radar to enhance perceptual robustness. While these technologies improve raw data acquisition, a critical gap persists in bridging perceptual inputs with contextual reasoning and actionable control outputs. 
Vision-language models attempt to mitigate this by grounding visual data in textual instructions, yet they often lack the situational awareness and adaptability required for complex and long-tail driving scenarios. 

Despite these technological advancements, autonomous driving systems continue to face limitations that hinder their ability to achieve human-like adaptability and safety in complex, dynamic environments. These primary limitations include: 
\begin{itemize}
    \item \textbf{Poor connectivity between functional modules} -- Isolated perception, language, and planning subsystems hinder cohesive scene understanding, reducing contextual coherence and safety in dynamic driving scenarios like merging or vehicle-following.
    \item \textbf{Lack of structured semantic understanding in perception data} -- Raw sensor data (e.g., LiDAR, images) lacks inherent meaning, requiring complex multi-stage processing (e.g., object detection, semantic segmentation, reasoning), which compromises reliability and interpretability in safety-critical decisions.
    \item \textbf{Limited generalization to unseen scenarios} --Rule-based or narrowly trained models struggle with unfamiliar scenarios, such as construction zones or erratic pedestrian behavior, lacking the robust, human-like reasoning needed for diverse conditions.
\end{itemize}
\begin{figure*}[t]
    \centering
    \includegraphics[width=0.95\textwidth]{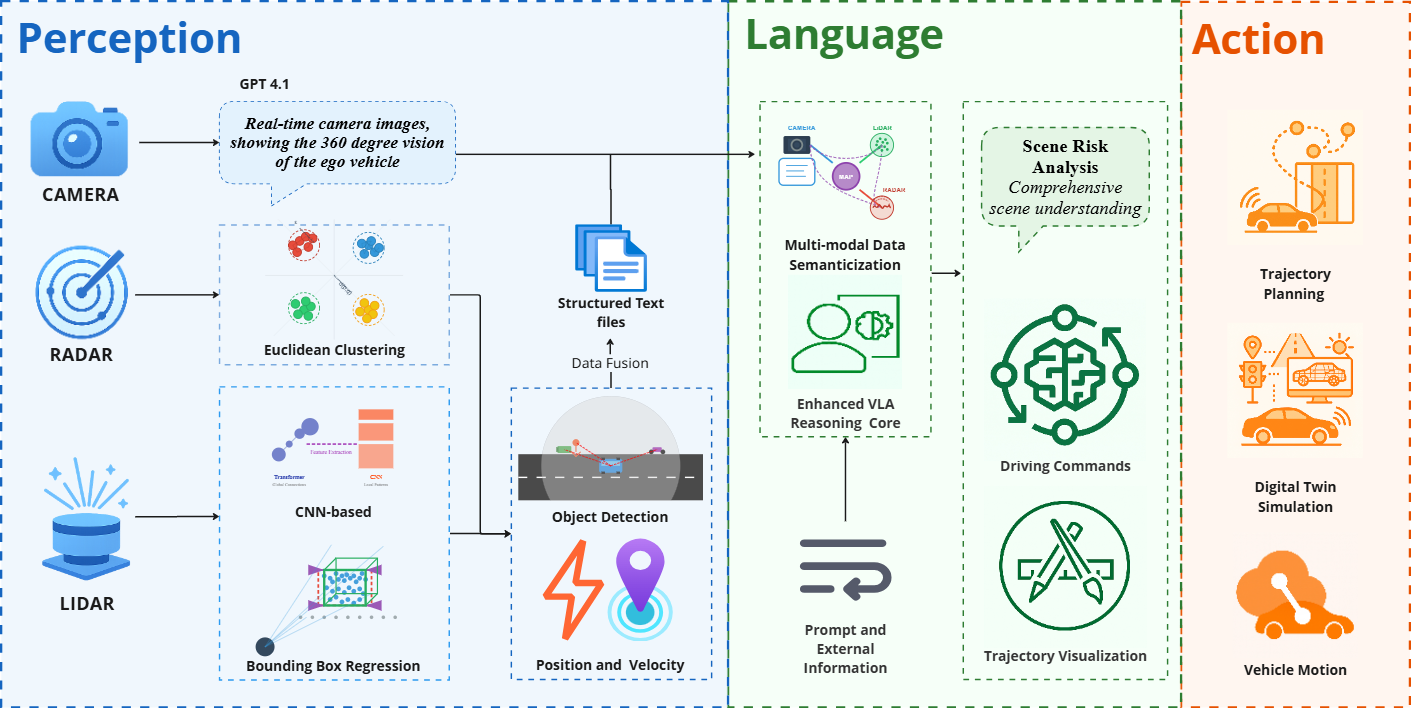} % Adjust width as needed
    \caption{Detailed workflow of the proposed framework for complex scene interpretation and motion control in autonomous driving: (1) Perception-layer, (2) Language-layer, and (3) Action-layer.}
    \label{fig:workflow_details}
\end{figure*}

However, large language models (LLMs) demonstrate exceptional capabilities in reasoning, planning, and interactive understanding. So to address these challenges, we propose a unified framework that integrates multi-sensor fusion with an LLM-augmented vision-language-action (VLA) agent. This framework enables autonomous systems to combine low-level perception with high-level cognitive reasoning, achieving explainable, adaptive, and safety-bounded decision-making in open-world driving environments. Our key contributions are outlined below:
\begin{itemize}
\item \textbf{Integrated Cognitive Framework} – We introduce a Perception-Language-Action (PLA) framework that tightly couples multi-modal perception with LLM-based reasoning and motion planning, enabling coherent and adaptive decision-making in complex urban environments.
\item \textbf{Multi-Sensor Semantic Fusion} – We develop a robust fusion module that combines LiDAR, radar, and camera data into structured scene descriptions, enhancing both spatial accuracy and semantic richness for downstream reasoning.
\item \textbf{Enhanced Generalization through Contextual Reasoning} – By incorporating LLM-driven reasoning, our framework improves generalization to unseen scenarios,
such as construction zones or unpredictable pedestrian behaviors, enabling robust
decision-making in complex environments.
%\item \textbf{Digital Twin Validation} – We propose a novel validation method using digital twin simulations based on real-world data, supporting explainable scenario reproduction and robust performance assessment in edge cases.
\item \textbf{Empirical Validation in Complex Urban Scenarios}: We demonstrate the effectiveness and real-time adaptability of our framework through a case study on the nuScenes dataset, focusing on a challenging urban intersection with a construction zone, achieving low prediction errors and robust navigation performance.
\end{itemize}

The rest of the paper is structured as follows. Section II reviews related work, discussing advances in multi-sensor fusion, vision-language models, and large language models for autonomous driving. Section III introduces the Perception-Language-Action framework, including multi-sensor fusion and vision-language-action architecture. Section IV details the experimental setup and a case study on an urban intersection scenario. Section V presents the results, and Section VI concludes with findings and future work.

\section{Related Work}
\label{sec:related_work}
The development of autonomous driving systems has spurred significant research across perception, decision-making, and control, with notable advances in multi-sensor fusion, vision-language integration, and reasoning-augmented architectures. This section surveys these efforts, emphasizing their strengths and limitations in addressing modular compartmentalization, data scarcity, and contextual rigidity. 

\subsection{Multi-Sensor Fusion for Perception}
Multi-sensor fusion has emerged as a cornerstone for robust perception in autonomous driving. Techniques integrating LiDAR, cameras, and radar enhance spatial and temporal scene understanding. For instance, \textit{PointPainting}~\cite{vora2020pointpainting} fuses LiDAR with camera-derived semantic features for improved 3D object detection, while \textit{TransFuser}~\cite{prakash2021multi} employs transformers to align multimodal features for trajectory prediction. Radar’s resilience to adverse weather has been leveraged in fusion frameworks like \textit{RadarNet}~\cite{yang2020radarnet} to complement vision-based systems. However, these approaches often treat perception as an isolated module, leading to potential error propagation in dynamic scenarios, such as occlusions or sudden obstacles.

\subsection{Vision-Language Models for Scene Understanding}
Vision-Language Models (VLMs) have gained traction for grounding visual perception in natural language, offering a bridge between sensory inputs and decision-making. Models like CLIP~\cite{radford2021clip} and ViLBERT~\cite{lu2019vilbert} enable semantic alignment between images and text. In autonomous driving, \textit{DriveVLM}~\cite{tian2024drivevlm} explores VLMs to interpret driving scenes via textual prompts and vision inputs, while \textit{VLMaps}~\cite{huang2023vlmaps} employs language-guided spatial reasoning for navigation. However, these models often struggle with real-time adaptability and fine-grained situational awareness, such as anticipating pedestrian intent or interpreting ambiguous traffic signals.

\subsection{Large Language Models in Decision-Making}
Large Language Models (LLMs) like GPT-4~\cite{openai2023gpt4} and LLaMA~\cite{touvron2023llama} have demonstrated potential in reasoning and planning for autonomous systems. 
\textit{GPT-Driver}~\cite{yang2023gptdriver} utilizes in-context learning to generate explainable driving trajectories, while \textit{DriveLLM}~\cite{Cui2024drivellm} aligns multimodal motion prediction with language prompts. 
Despite these advances, LLM-based approaches often operate in isolation from low-level perception and control loops, limiting their ability to adapt to real-time stochastic conditions such as urban intersections or adverse weather ~\cite{yang2023llm4drive}.

\subsection{Integrated Architectures for Autonomous Driving}
Efforts to unify perception, reasoning, and action have led to hybrid architectures that blend modular and end-to-end learning approaches. Transformer-based frameworks like  \textit{UniAD}~\cite{hu2023planning} integrate perception, prediction, and planning into a unified pipeline, reducing error propagation while maintaining interpretability. \textit{Reason2Drive}~\cite{nie2024reason2drive} and \textit{RDA-Driver}~\cite{huang2024making} introduce reasoning-based decision-making but remain constrained by a lack of joint optimization with perception modules. Meanwhile, \textit{Text2motion}~\cite{lin2023text2motion} integrates neural motion planners with language interfaces, underscoring the growing need for explainability and adaptability in autonomous systems.

%\subsection{Datasets and Benchmarking Challenges}
%The performance of autonomous driving models is heavily influenced by available datasets, such as nuScenes~\cite{caesar2020nuscenes} and Waymo Open~\cite{sun2020waymo}, which provide rich multi-sensor annotations but are biased toward common scenarios. Synthetic datasets like CARLA~\cite{dosovitskiy2017carla} enable scalable scenario generation but face sim-to-real transfer challenges. Recent benchmarks such as \textit{DriveLM}~\cite{sima2024drivelm} introduce language-annotated driving scenarios but lack actionable control signals, limiting their utility for training Vision-Language-Action (VLA) models. Data diversity and alignment across modalities remain critical bottlenecks in developing robust, generalizable autonomous systems.

%\subsection{Challenges and Future Directions}
%Despite significant progress, gaps remain in achieving real-time adaptability and contextual reasoning in autonomous systems. The fragmentation between perception, reasoning, and action continues to limit responsiveness, while data scarcity constrains model generalization to edge cases. Additionally, the interpretability of end-to-end learning frameworks remains an open challenge. Future research should focus on integrating multi-sensor fusion with LLM-augmented VLA architectures, leveraging language-guided control mechanisms for safer and more explainable autonomy in complex driving environments.

\section{Methodology and Workflow}

In this section, we introduce a framework to tackle the challenges of poor functional module connectivity, limited generalization to unseen scenarios, and evaluation difficulties in autonomous driving systems. Our approach integrates multi-sensor fusion with a large language model-augmented Vision-Language-Action (VLA) architecture to enhance system performance and adaptability.

\subsection{Overview of the Proposed Framework}
The pipeline of our framework enables advanced scene interpretation and motion control for dynamic driving scenarios, as depicted in Figure~\ref{fig:workflow_details}. The pipeline comprises three primary layers:

\begin{itemize}
    \item \textbf{Perception Layer}:
     Raw sensor data from cameras, radar, and LiDAR are processed for a cohesive environmental representation. 360-degree camera images are interpreted using advanced models like GPT-4.1for interpretation. Concurrently, radar data undergoes Euclidean clustering for object delineation based on spatial proximity. LiDAR point clouds are processed via CNNs with bounding box regression for 3D object detection. A data fusion mechanism then integrates LiDAR and Radar outputs, producing structured text files that contain precise position and velocity information for detected objects. This fusion process leverages the complementary strengths of each modality to enhance the accuracy of object detection and multi-target tracking.
    \item \textbf{Language Layer}:
    The language layer processes structured text files and camera images, converting fused perception data into semantically rich representations. An enhanced Vision-Language-Action (VLA) Reasoning Core conducts comprehensive scene risk analysis and understanding, enabling context-aware decision-making. Furthermore, the system integrates prompt-based inputs and external information (e.g., real-time traffic alerts, prior experience) to enrich the contextual understanding. Based on this comprehensive analysis, the language layer generates precise driving commands and visualizes the planned trajectories.
    \item \textbf{Action Layer}:
    The action layer receives driving commands and trajectory visualizations from the language layer. It handles detailed trajectory planning, converting high-level commands into precise, actionable vehicle paths. These paths are validated using high-fidelity digital twin simulations, which replicate real-world scenarios to ensure safety and efficiency. The layer’s final output is direct control of vehicle motion, guiding the autonomous system effectively in dynamic driving environments.
\end{itemize}

\begin{figure}[t]
    \centering
    \includegraphics[width=0.95\columnwidth]{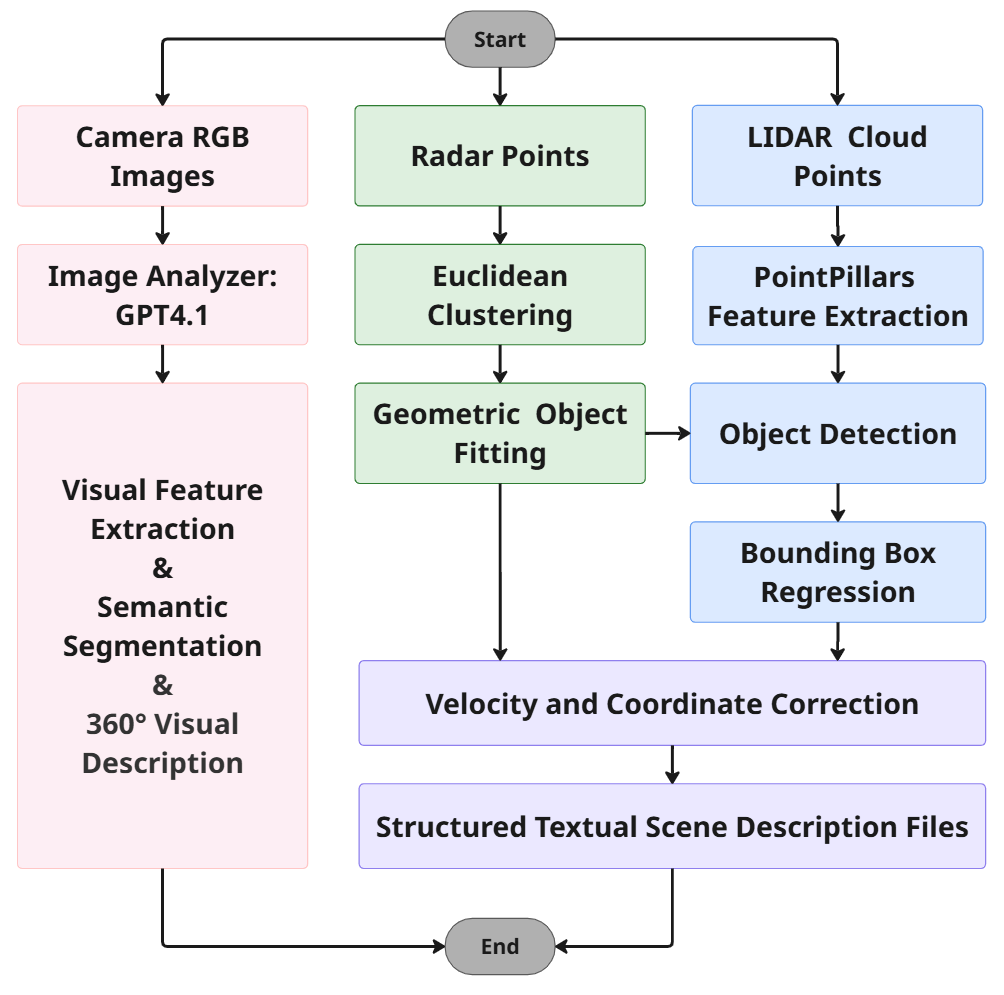}
    \caption{Multimodal sensor fusion flow}
    \label{fig:sensor_fusion_flow}
\end{figure}
\subsection{Multi-Sensor Fusion Module}
% Details of sensor processing and fusion (e.g., how LiDAR, cameras, radar are combined)
%The multi-sensor fusion module forms the foundational perception layer of our framework, integrating data from three primary sensors—camera, LiDAR, and radar—to achieve a comprehensive and interpretable understanding of the driving scene, which informs the language layer for scene understanding and decision-making. 
The multi-sensor fusion module integrates LiDAR, and radar data to create a structured representation of the ego vehicle’s state and surrounding obstacles within 50 meters, enabling robust scene understanding, object tracking, and trajectory planning. Cameras provide visual data for feature extraction and semantic segmentation, LiDAR offers 3D point clouds for geometric understanding, and radar ensures reliable velocity estimation in adverse conditions. The process, shown in Figure~\ref{fig:sensor_fusion_flow}, uses parallel pipelines to combine data for downstream tasks.
%This module fuses multimodal sensor data from cameras, LiDAR, and radar to create a structured representation of the ego vehicle’s state and its surrounding obstacles within a 50-meter radius, enabling robust object tracking, and semantic grounding for downstream tasks like scene analysis and trajectory planning. Cameras provide high-resolution RGB images for visual feature extraction and semantic segmentation (e.g., lane markings, traffic signs), LiDAR delivers precise 3D point clouds for accurate depth and geometric understanding (e.g., obstacle detection, distance measurement), and radar ensures reliable velocity estimation and robustness even in adverse conditions (e.g., rain, fog) for tracking dynamic objects like vehicles and pedestrians.

\begin{table}[!t]
    \caption{An Example of Ego Vehicle State and Relative Obstacles in a structured textual scene description file}
    \label{tab:scene_description}
    \centering
    \footnotesize
    \begin{tabular}{ll}
        \toprule
        \textbf{Ego Vehicle Information} & \\
        \midrule
        Label & ego\_vehicle \\
        Dimension (m) & (3.99, 2.06, 1.84) \\
        Position (m) & (0, 0, 0) \\
        Distance (m) & 0 \\
        Velocity (m/s) & $v_x = 8.28$, $v_y \approx 0$, $v_z = 0$\\
        Speed (m/s) & 8.28 \\
        \midrule
        \textbf{Obstacle Information} & \\
        \midrule
        \textbf{Obstacle 1} & \\
        Label & human.pedestrian.adult \\
        Partition & Front-right \\
        Position (m) & (25.17, -21.64, 0.86) \\
        Distance (m) & 33.20 \\
        Velocity (m/s) & $v_x = 1.26$, $v_y = -0.06$, $v_z = -0.03$ \\
        Speed (m/s) & 1.26 \\
        \midrule
        \textbf{Obstacle 2} & \\
        Label & vehicle.truck \\
        ... & ... \\
        %Position (m) & (3.19, 4.46, 1.62) \\
        %Distance (m) & 5.72 \\
        %Velocity (m/s) & $v_x = 0.23$, $v_y = 0.03$, $v_z = 0.00$ \\
        % Speed (m/s) & 0.23 \\
        \bottomrule
    \end{tabular}
\end{table}
%Figure~\ref{fig:sensor_fusion_flow} presents the overall sensor fusion procedure, which integrates information from multiple modalities, including cameras, radar, and LiDAR sensors, to achieve robust and comprehensive scene understanding. The system employs parallel data processing pipelines for each sensor, followed by a fusion stage that combines the extracted information for downstream applications.
The camera pipeline processes RGB images using an advanced AI-based module (ChatGPT-4.1) for visual feature extraction, semantic segmentation, and generating a 360-degree visual description, identifying elements such as traffic signs, pedestrians, and lane boundaries. LiDAR point clouds are processed using the PointPillars~\cite{lang2019pointpillars} architecture for feature extraction, followed by object detection and bounding box regression to localize and estimate object dimensions in 3D space. Radar data, provided as point measurements, undergoes Euclidean clustering to form object-level groupings, followed by geometric object fitting. Radar and LiDAR pipelines are integrated in a velocity and coordinate correction block to ensure temporal and spatial consistency. 
%The camera stream begins with the acquisition of RGB images, which are analyzed using an advanced AI-based module (ChatGPT-4.1). This module performs visual feature extraction, semantic segmentation, and produces a 360-degree visual description, identifying elements such as traffic signs, pedestrians, and lane boundaries.
%LiDAR point clouds are processed using the PointPillars~\cite{lang2019pointpillars} deep learning architecture for feature extraction. The subsequent object detection and bounding box regression modules localize objects and estimate their dimensions within the three-dimensional space.
%Radar data, provided as point measurements, is subjected to Euclidean clustering to form object-level groupings. Geometric object fitting is then applied to these clusters, yielding geometric parameters of the detected objects from Lidar. 

The fused scene information is  processed into structured text files, detailing a scene relative to an ego vehicle, comprising two sections:
\begin{itemize}
    \item \textbf{Ego Vehicle Information}: Attributes include the label (object type), dimensions (\si{\meter} for length, width, height), position \((x, y, z)\) in the ego-vehicle coordinate system (\si{\meter}), distance to origin (\si{\meter}), velocity \((v_x, v_y, v_z)\) (\si{\meter\per\second}), and scalar speed (\si{\meter\per\second}).
    \item \textbf{Obstacles}: A list of objects within a \SI{50}{\meter} radius of the ego vehicle, each with attributes: label (obstacle type, e.g., pedestrian, vehicle, barrier, or unknown), partition (relative position, e.g., front, back, right), position \((x, y, z)\) (\si{\meter}), distance (\si{\meter}), velocity \((v_x, v_y, v_z)\) (\si{\meter\per\second}), and scalar speed (\si{\meter\per\second}).
\end{itemize}

Table~\ref{tab:scene_description} illustrates examples of these outputs, detailing the ego vehicle’s state and the attributes of surrounding obstacles.

\subsection{Augmented Vision-Language-Action (VLA) Architecture}
We propose an integrated Augmented Vision-Language-Action (VLA) architecture that unifies perception, scene analysis, and action planning for autonomous navigation. Leveraging large language models (LLMs) as core reasoning engines, our approach enables intuitive interpretation of complex driving scenarios, thereby enhancing explainability and robust handling of ambiguous cases that challenge conventional modular systems.

\begin{figure}[t]
    \centering
    \includegraphics[width=0.95\columnwidth]{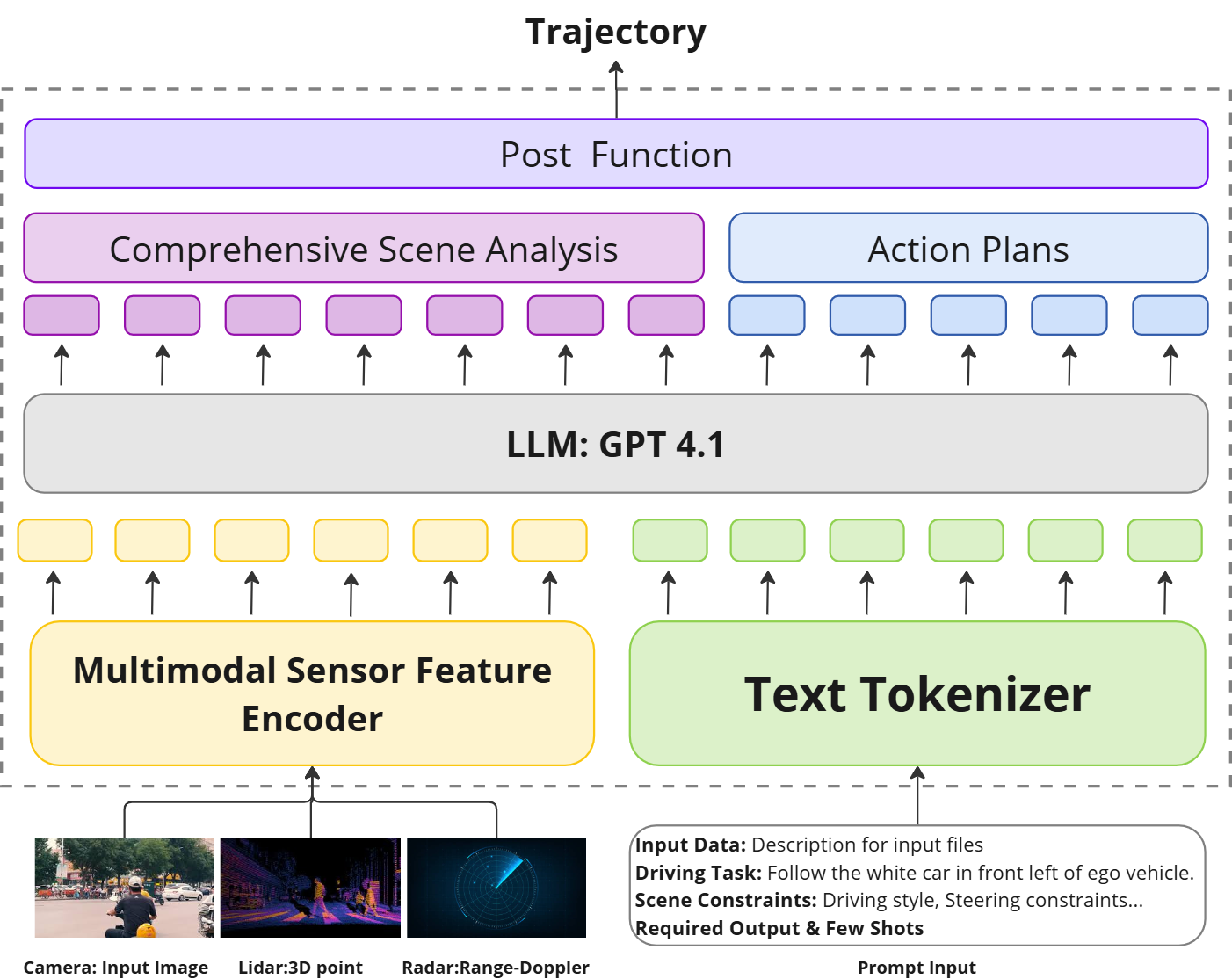}
    \caption{The architecture of Augmented Vision-Language-Action model.}
    \label{fig:Augmented_VLA_model}
\end{figure}

As illustrated in Fig.~\ref{fig:Augmented_VLA_model}, the architecture accepts multimodal sensor streams—camera images, LiDAR point clouds, and radar data—which are processed by a Multimodal Sensor Feature Encoder. This encoder extracts high-level representations from heterogeneous sensor modalities. Simultaneously, natural language inputs, including environmental conditions, driving tasks, and scene constraints, are processed via a dedicated Text Tokenizer.

Both feature streams are integrated and fed into an LLM (GPT 4.1), which serves as the central reasoning component. The LLM generates two main outputs: Comprehensive Scene Analysis, providing environmental interpretation, and Action Plans, defining navigation strategy. These outputs are further post-processed to produce the final vehicle trajectory.

This architecture unifies perception, language-based reasoning, and action planning within a single framework, thereby advancing the capabilities of autonomous systems toward more interpretable and adaptable behavior.

\section{Case Study: Urban Intersection Following with Construction Zone}
In this section, we present the experimental evaluation of our multi-layered framework for autonomous driving in complex environments. Although various scenarios were tested, this paper focuses on a representative case study: a following task at an urban intersection with an active construction zone. This challenging scenario enables a targeted assessment of the Perception-Language-Action (PLA) architecture’s capabilities in advanced scene understanding and robust decision-making. While the results demonstrate the effectiveness of our framework, further validation across more diverse scenarios will be conducted in future work.
\begin{figure}[!t]
    \centering
    \begin{subfigure}[c]{0.49\linewidth}
        \centering
        \includegraphics[width=\linewidth]{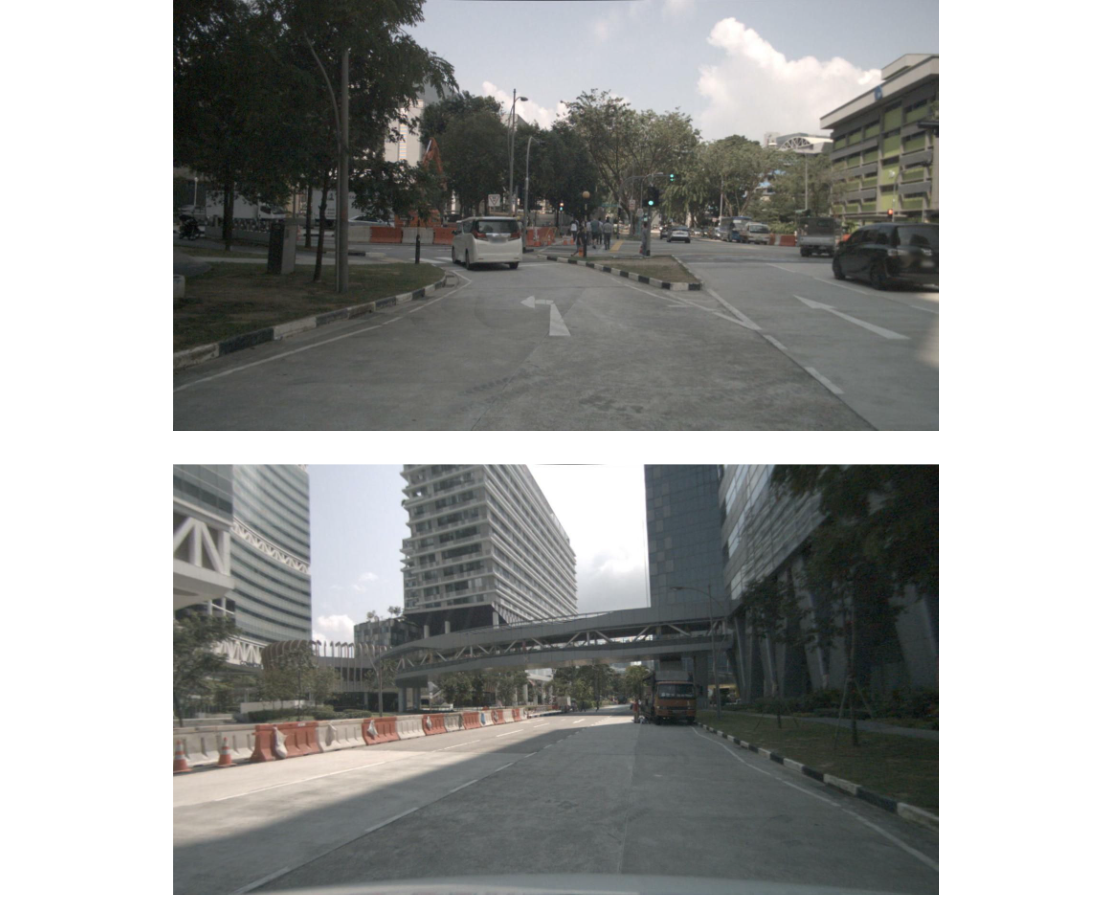}
        \caption{Front and rear camera views of the intersection with construction zones.}
        \label{fig:front_view}
    \end{subfigure}
    \hfill
    \begin{subfigure}[c]{0.49\linewidth}
        \centering
        \includegraphics[width=\linewidth]{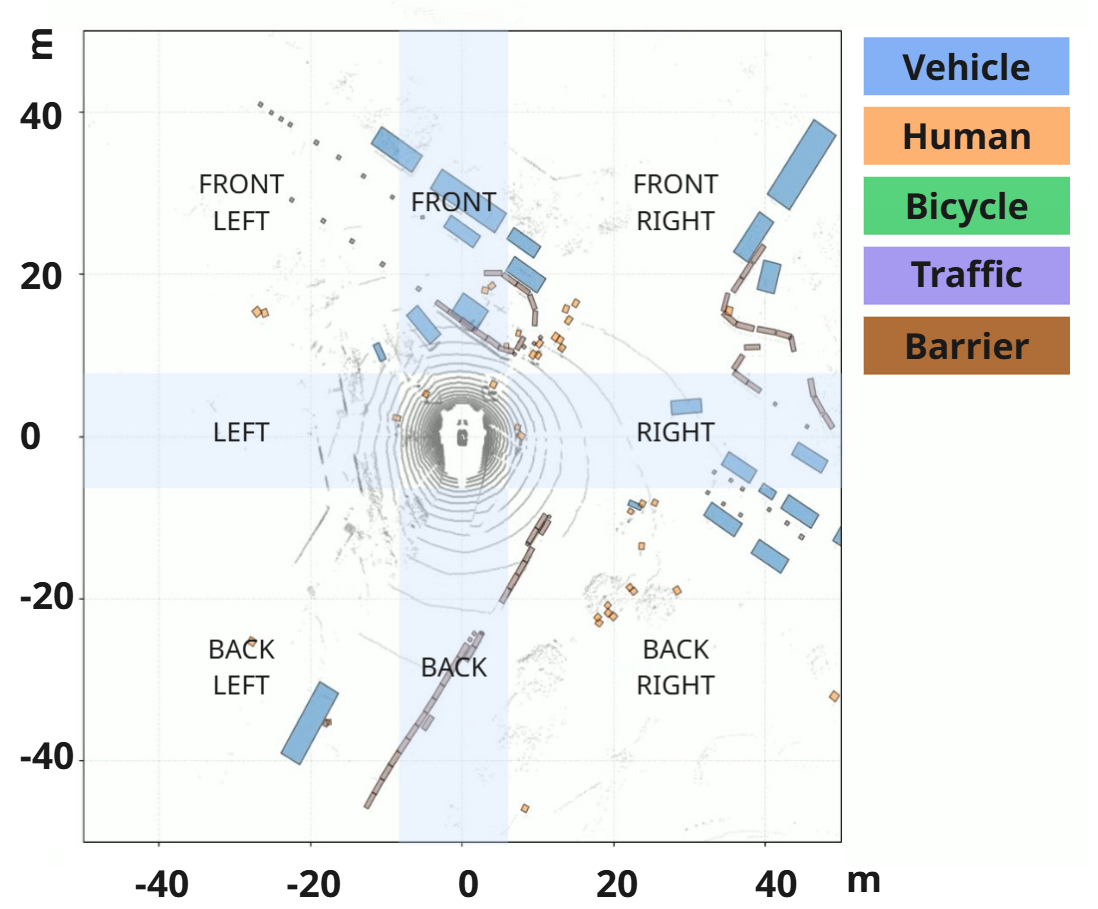}
        \caption{Bird's-eye view visualization of the intersection with construction zones.}
        \label{fig:sensor_fusion_BV_visualization}
    \end{subfigure}
    \caption{Illustration of the selected urban intersection scenario.}
    \label{fig:scenario_views}
\end{figure}
\begin{table}[!t]
    \caption{Key Performance Metrics}
    \label{tab:key_metrics}
    \centering
    \resizebox{0.95\linewidth}{!}{
    \begin{tabular}{lcc}
    \toprule
    \textbf{Category} & \textbf{Metric} & \textbf{Value} \\
    \midrule
    Speed Prediction & MAE (m/s) & 0.39 \\
     & R$^2$ Score & 0.923 \\
    \midrule
    Steering Angle Prediction & MAE ($^\circ$) & 2.52 \\
     & R$^2$ Score & 0.537 \\
    \midrule
    Trajectory & ADE (m) & 1.013 \\
     & FDE (m) & 2.026 \\
    \bottomrule
    \end{tabular}
    }
\end{table}
\begin{figure}[!t]
    \centering
    \begin{subfigure}[b]{0.9\linewidth}
        \centering
        \includegraphics[width=\linewidth]{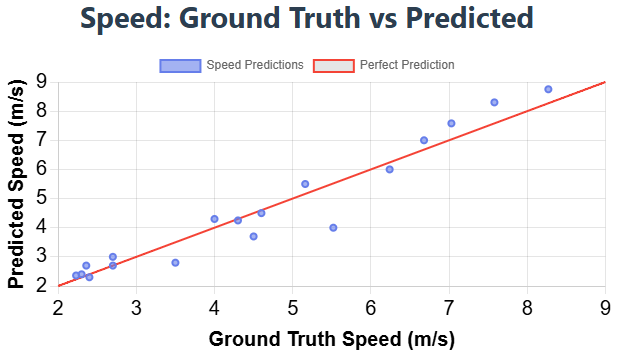}
        \caption{Speed: ground truth vs predicted. The red line indicates perfect prediction.}
        \label{fig:speed_result}
    \end{subfigure}
    \hfill
    \begin{subfigure}[b]{0.9\linewidth}
        \centering
        \includegraphics[width=\linewidth]{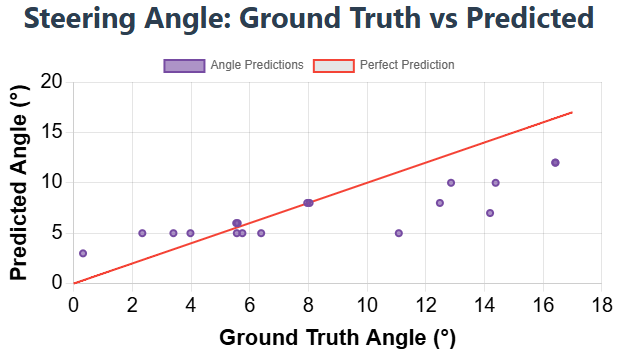}
        \caption{Steering angle: ground truth vs predicted. The red line indicates perfect prediction.}
        \label{fig:steering_angle_result}
    \end{subfigure}
    %\vskip\baselineskip
    %\begin{subfigure}[b]{\linewidth}
    %    \centering
    %    \includegraphics[width=\linewidth]{tum/speed_steer.png}
    %    \caption{Speed and steering trajectory: predicted vs ground truth. The green solid line represents the ground truth trajectory, while the orange dashed line shows the predicted trajectory. Green and red squares indicate the start and end points respectively.}
    %    \label{fig:speed_steer_result}
    %\end{subfigure}
    \caption{Comparison of model predictions with ground truth values for speed and steering angle.}
    \label{fig:prediction_results}
\end{figure}
\subsection{Scenario Description}
The selected case study focuses on a following task at an urban intersection with an active construction zone. Here, the ego vehicle navigates a signalized intersection where barriers and warning signs cause partial lane occlusion, temporary lane shifts, and disrupted traffic flow. The environment is further complicated by irregular vehicle movements, dynamic obstacles (e.g., workers and equipment), unpredictable surrounding behavior, rapidly changing road conditions, and occlusions from construction machinery. This scenario was selected for its high unpredictability and complexity, posing significant challenges to perception, planning, and decision-making. It serves as a rigorous testbed to evaluate the robustness and adaptability of our framework under realistic, non-ideal conditions.

Figure~\ref{fig:scenario_views} illustrates the selected urban intersection scenario with an active construction zone. Specifically, Figure~\ref{fig:front_view} shows the front and rear camera perspectives, and Figure~\ref{fig:sensor_fusion_BV_visualization} provides a bird's-eye view visualization based on sensor fusion data.

\subsection{Experimental Setup}

\subsubsection{Dataset}
We used the nuScenes dataset as the primary data source for our experiments. nuScenes is a large-scale, publicly available autonomous driving dataset that provides synchronized data from multiple sensors, including cameras, LiDAR, radar, and GPS/IMU, collected in urban environments. 

%in an unseen intersection with construction zones, our framework processes LiDAR, camera, and radar data to detect the environment. Janus Pro identifies construction barriers and signage from camera images, while Euclidean clustering on LiDAR point clouds separates static obstacles (e.g., cones) from dynamic entities (e.g., pedestrians). Radar provides velocity data despite dust or obstructions. Geometric data fusion creates a labeled scene representation, which the Language layer enhances with road closure alerts and traffic data. The VLA reasoning core analyzes the intersection’s layout, assesses risks (e.g., pedestrian crossing), and decides on actions like detours or speed reduction. The Action layer generates safe trajectories and motion commands (e.g., 10 km/h reduction, adaptive braking), validated via CARLA digital twin simulations for dynamic adaptability, showcasing robust multi-sensor fusion and LLM-driven decision-making in complex environments.
\subsubsection{Metrics}
We evaluate system performance using standard metrics for autonomous driving: mean absolute error (MAE)\cite{willmott2005advantages} and coefficient of determination ($R^2$ score)\cite{nagelkerke1991note} for speed and steering angle prediction, as well as average displacement error (ADE) and final displacement error (FDE) for trajectory accuracy~\cite{alahi2016social, caesar2020nuscenes}. These metrics respectively capture absolute prediction error, regression fit, average trajectory deviation, and endpoint accuracy. They provide a comprehensive assessment of the system’s ability to estimate motion parameters and maintain accurate trajectories in complex urban scenarios.
\subsubsection{Procedure}
For the scenes in the dataset, we sample various frames to obtain the original data. The proposed PLA framework is then applied to each sampled frame to predict the future trajectory for the next one second.  
And to emulate real-world perception-to-decision workflows, we design a structured prompt that guides GPT-4.1 to analyze traffic scenes and generate driving strategies. The input includes six surrounding camera views, a front-facing image for trajectory overlay, and a structured file with ego status and obstacle information.
The prompt specifies the driving task, lane information, safe lateral deviation in a lane(within $\pm$1.0\,m), and typical steering rate (5–15$^\circ$/s). The model is instructed to assess risks and output:
\begin{itemize}
    \item \textbf{Driving Commands:} speed action (accelerate/decelerate/maintain), steering direction (left/right) and angle for the next second;
    \item \textbf{Explanation:} reasoning based on perception and motion data.
\end{itemize}
%Finally, the predicted trajectories are compared with the corresponding ground truth trajectories, enabling quantitative evaluation of the model's prediction accuracy.
%By inputting the unified sensor fusion text and external information to the contextual fusion engine, we can get a josn file which detailed describes the environmental conditions, and by this file, we can easily get the textual scene analysis and  use this josn file to generate openScenario, by using the customised josn file and openScenerio files, we can generate Carla scenarios, and the Carla simulation will be discussed in Section ~\ref{sec:validation}. 

%There are two parts should be shown in this section, first, how the contextual fusion engine work and the result is, second is how to use the generated josn file to describe the scene comprehensively. 

%We evaluate performance using metrics specifically targeting complex scene interpretation and trajectory prediction: scene comprehension accuracy (identifying critical objects and their relationships), contextual reasoning precision (correctly interpreting situational hazards), and path prediction quality (measured through trajectory deviation and anticipatory response to dynamic obstacles). Our validation in CARLA simulations compares these metrics against conventional perception-only systems to quantify improvements in navigation through ambiguous and unstructured environments.
\begin{figure*}[t]
    \centering
    \subfloat[Steering angle heat map\label{fig:steer_heatmap}]{
        \includegraphics[width=0.3\linewidth]{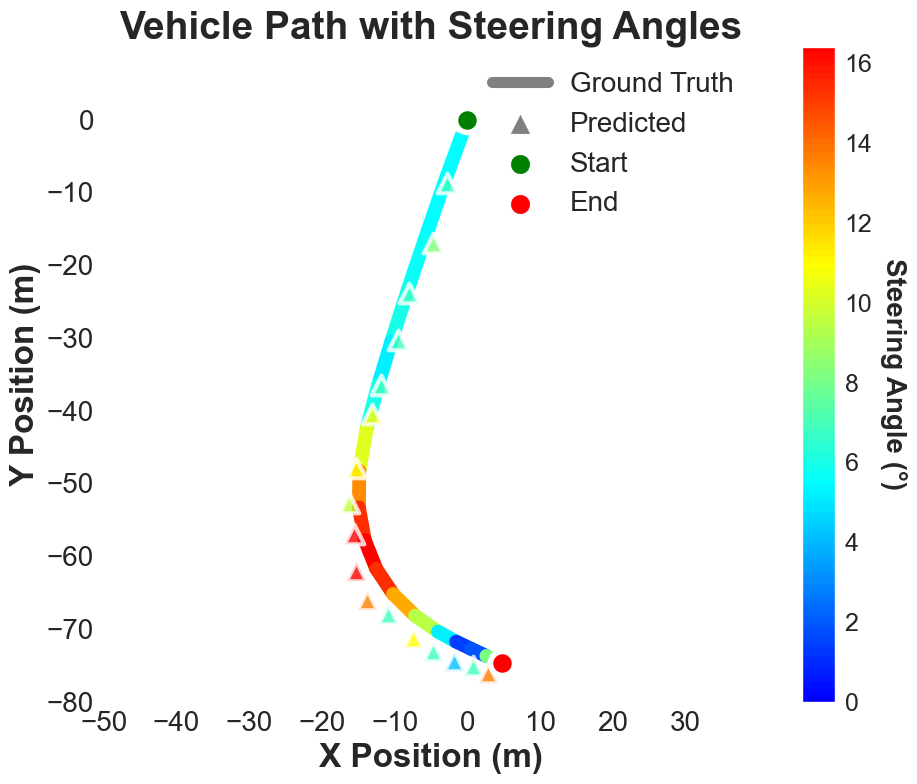}
    }
    \hfill
    \subfloat[ADE heat map\label{fig:ade_heatmap}]{
        \includegraphics[width=0.3\linewidth]{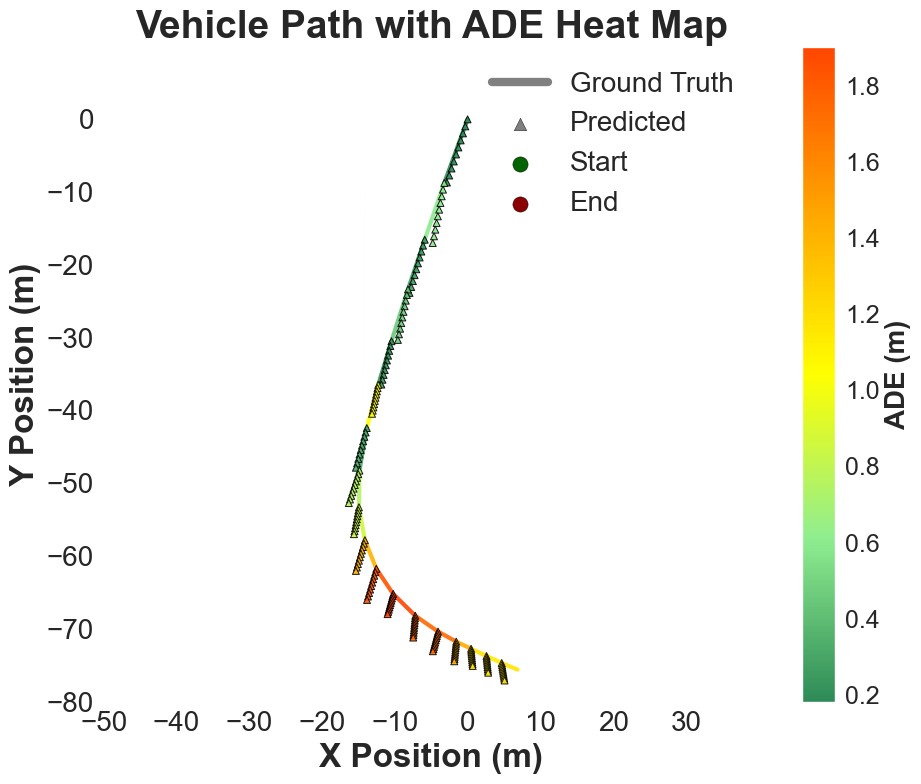}
    }
    \hfill
    \subfloat[FDE heat map\label{fig:fde_heatmap}]{
        \includegraphics[width=0.3\linewidth]{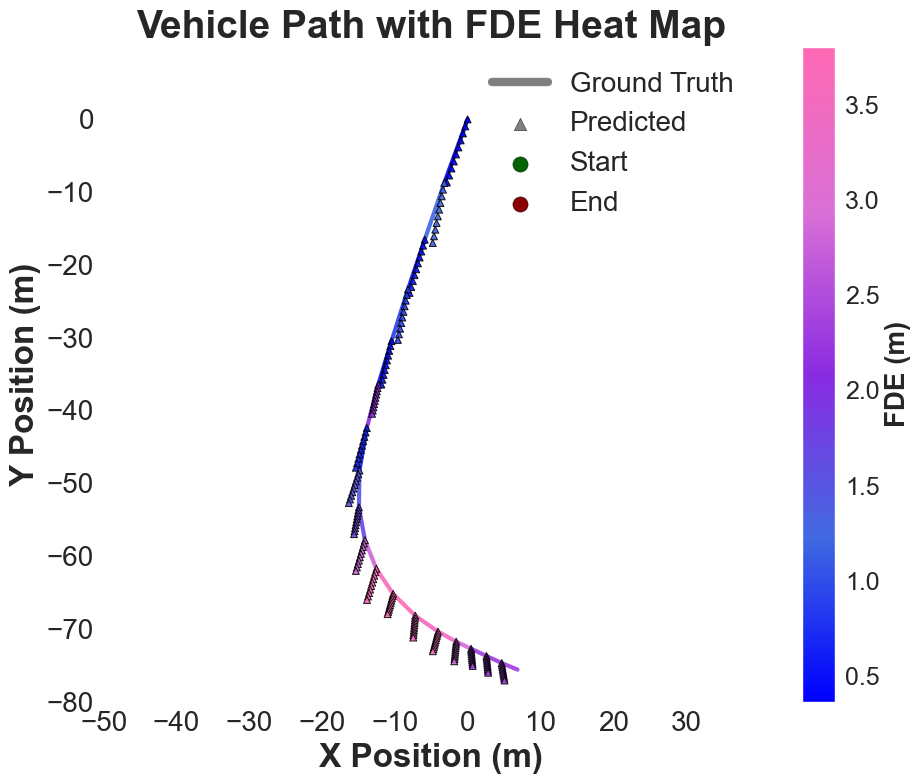}
    }
    \caption{Trajectory-based heat map visualizations for the same following task: (a) steering angle, (b) average displacement error (ADE), and (c) final displacement error (FDE). The color bars indicate the corresponding error or angle magnitude along the predicted path.}
    \label{fig:heatmap_viz}
\end{figure*}

\begin{figure}[!t]
    \centering
    \subfloat[Urban intersection entry\label{fig:scene5_predicted}]{
        \includegraphics[width=0.45\linewidth]{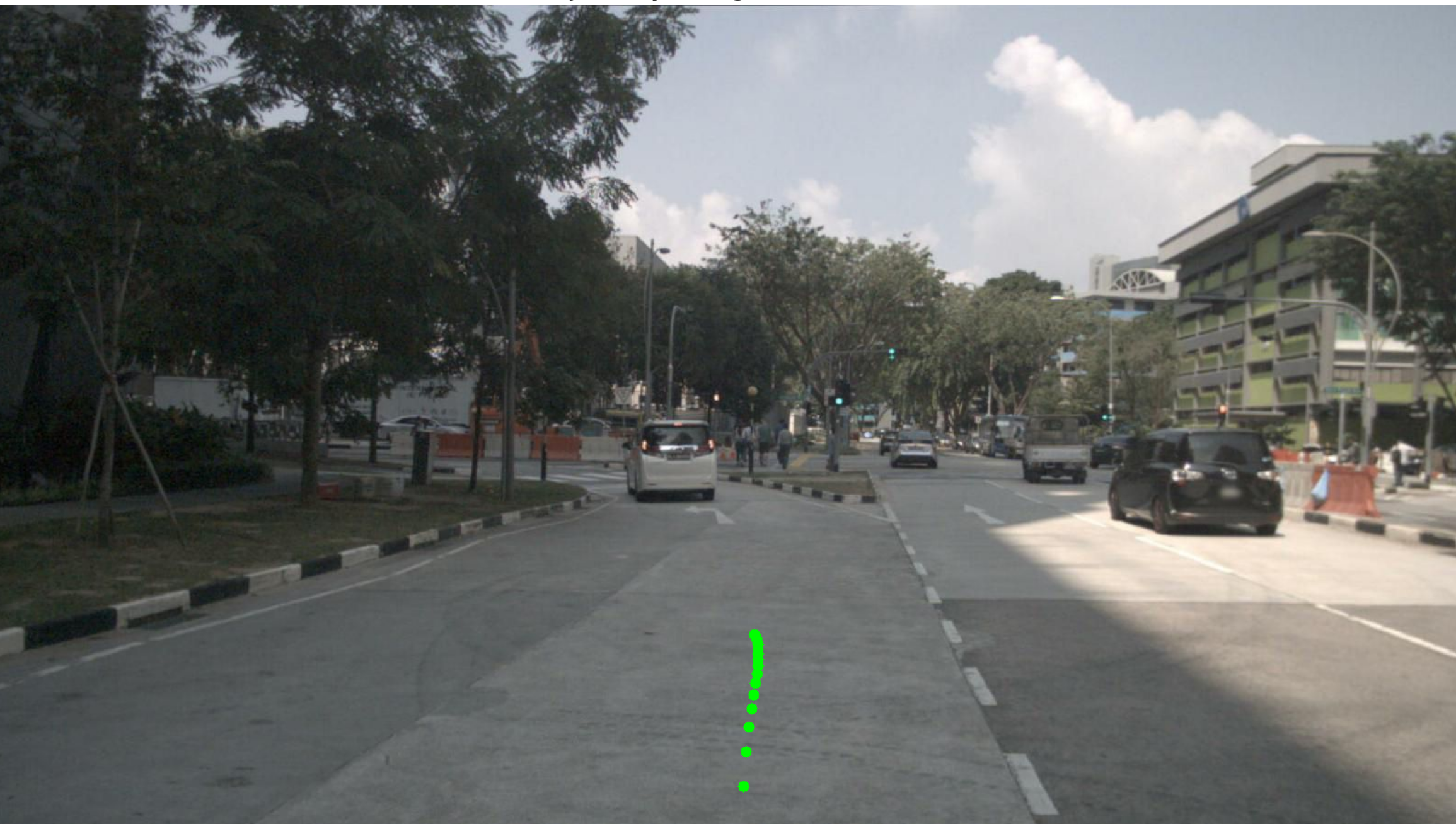}
    }
    \hfill
    \subfloat[Intersection with construction zone\label{fig:scene11_predicted}]{
        \includegraphics[width=0.45\linewidth]{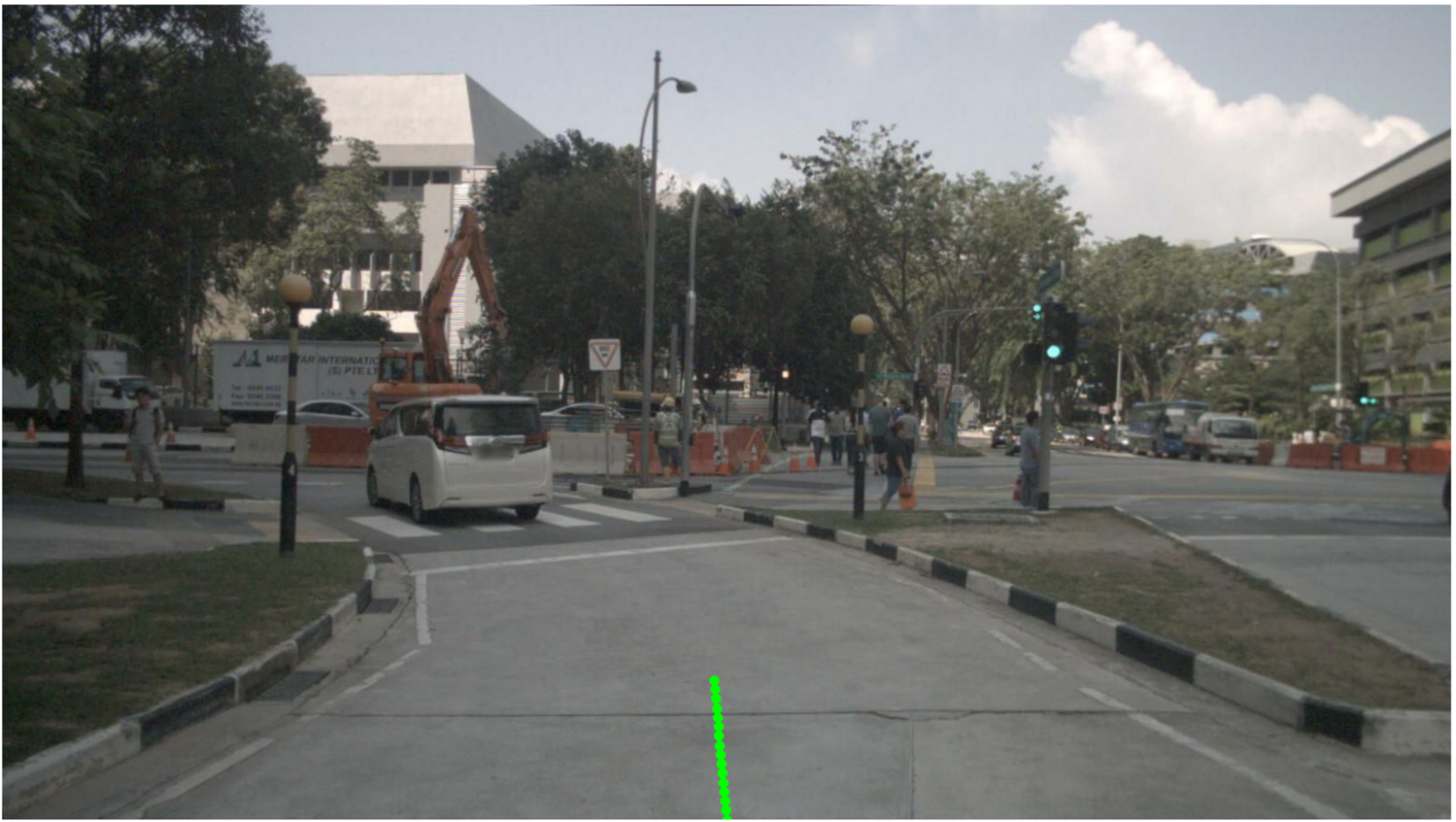}
    }
    \vfill
    \subfloat[Curved lane following\label{fig:scene21_predicted}]{
        \includegraphics[width=0.45\linewidth]{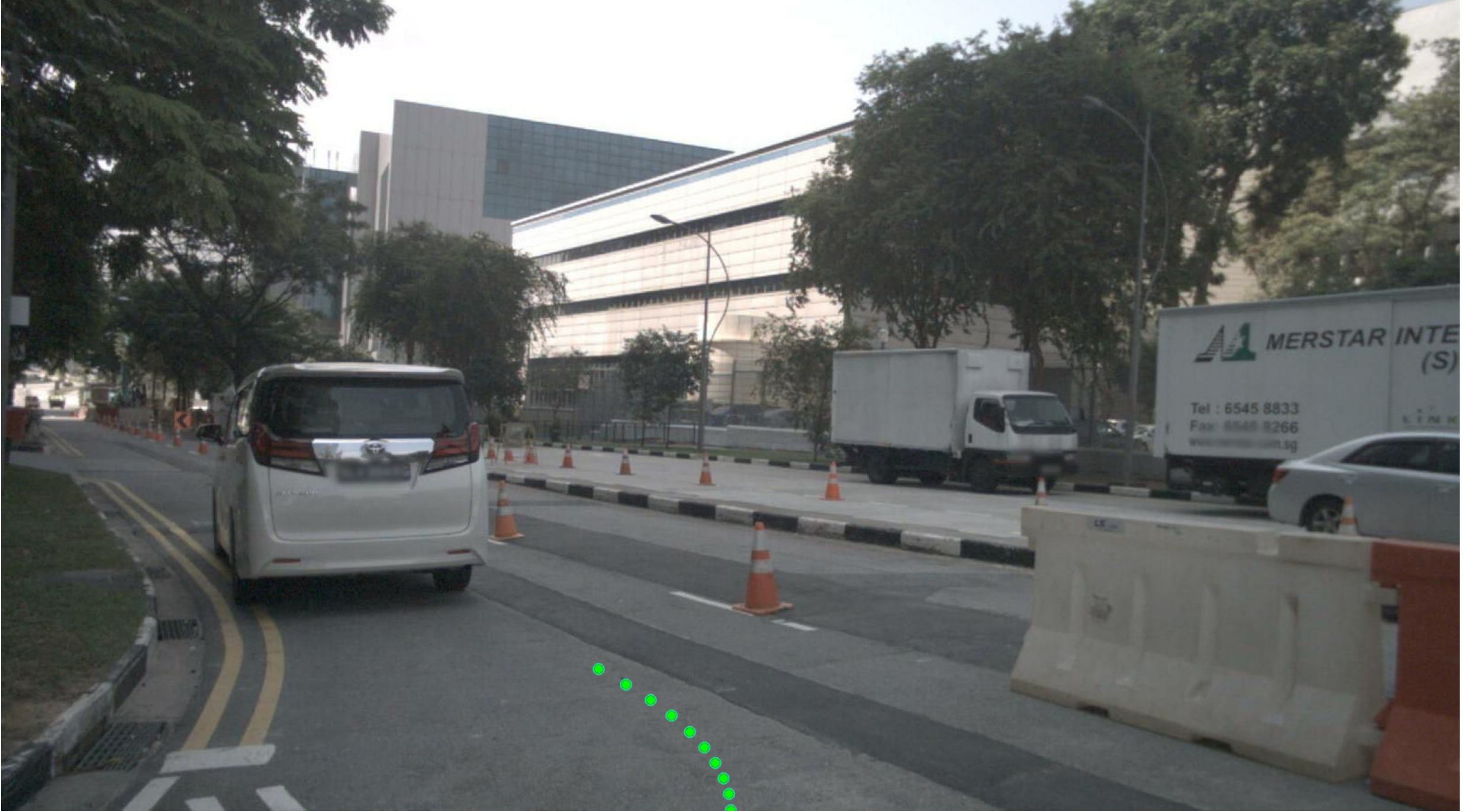}
    }
    \hfill
    \subfloat[Straight lane with turning white car\label{fig:scene35_predicted}]{
        \includegraphics[width=0.45\linewidth]{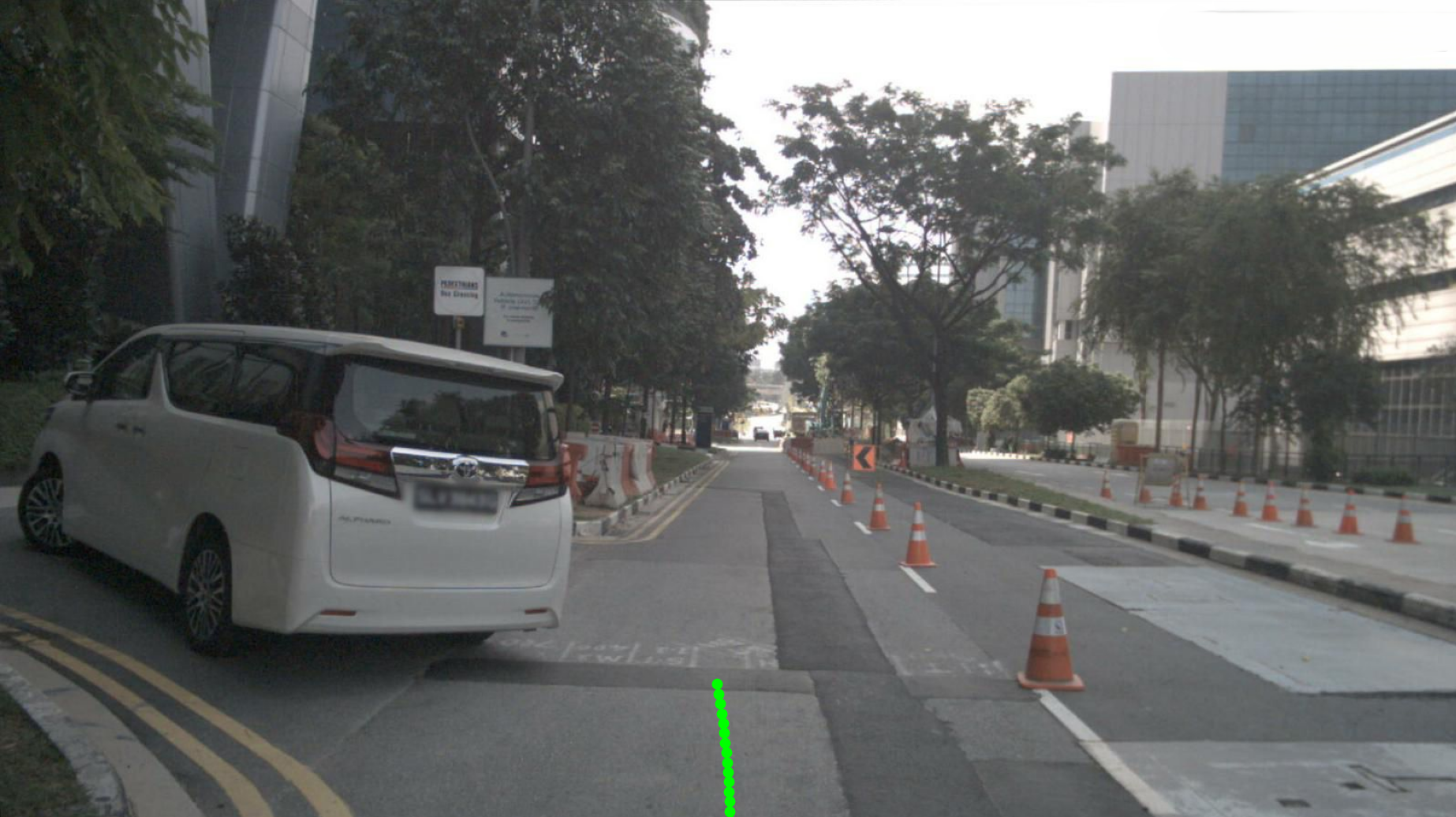}
    }
    \caption{Results of trajectory prediction by the PLA framework for the "following the front white car" task across diverse situations. Predicted trajectories are shown in green.}
    \label{fig:qualitative_plafw}
\end{figure}

\subsection{Results}

The qualitative and quantitative results are summarized in Table~\ref{tab:key_metrics} and Figures\ref{fig:speed_result},and ~\ref{fig:steering_angle_result}. 
For speed prediction, the model achieves a low mean absolute error (MAE) of 0.39~m/s and a high $R^2$ score of 0.923, indicating accurate and reliable performance across the test scenes. Most predicted speed values are closely aligned with the ground truth, as shown by the clustering along the diagonal in Figure~\ref{fig:speed_result}.

Steering angle prediction, however, is more challenging, with an MAE of 2.52$^\circ$ and an $R^2$ score of 0.537. Figure~\ref{fig:steering_angle_result} shows a wider spread of points at higher steering angles, where the model tends to slightly underestimate the actual values. These results indicate that while the model captures general trends, there remains room for improvement in precise steering control. 

%Figure~\ref{fig:speed_steer_result} compares the predicted and ground truth trajectories in the speed-steering angle coordinate system. The predicted trajectory (orange dashed line) generally follows the ground truth (green solid line), capturing overall trends but with some deviations in sharp turning scenarios.

For trajectory evaluation, the average displacement error (ADE) is 1.013~m and the final displacement error (FDE) is 2.026~m, demonstrating robust trajectory tracking overall, but highlighting the need for further improvement in steering accuracy. In summary, the results validate the effectiveness of the proposed model in speed and trajectory prediction, while identifying steering angle prediction under complex conditions as an area for future work.

% Metrics and testing strategy
\subsection{Discussion}
We present heat map visualizations in Fig.~\ref{fig:heatmap_viz} to assess prediction performance. The steering angle heat map (Fig.~\ref{fig:steer_heatmap}) shows predicted steering commands aligned with ground truth trajectories. The ADE (Fig.~\ref{fig:ade_heatmap}) and FDE (Fig.~\ref{fig:fde_heatmap}) heat maps illustrate spatial distributions of average and final displacement errors along predicted paths, respectively. These demonstrate low errors across most trajectory segments, validating robustness and accuracy. However, conservative steering angles during curves result in larger turning radii, causing slight deviations from inner lane alignment, though still safe for driving.

To further demonstrate the effectiveness of our PLA framework in real-world scenarios, Fig.~\ref{fig:qualitative_plafw} presents four representative frames from the “following the front white car” task across diverse urban settings. Predicted trajectories are shown in green. The selected scenes include: entering an urban intersection (Fig.~\ref{fig:scene5_predicted}), navigating through a construction zone (Fig.~\ref{fig:scene11_predicted}), curved lane following (Fig.~\ref{fig:scene21_predicted}), and approaching a turning vehicle in a straight lane (Fig.~\ref{fig:scene35_predicted}). In all cases, predictions closely follow the lead vehicle’s path, demonstrating that the PLA framework effectively adapts to complex layouts and dynamic conditions. These qualitative results reinforce its robustness and practical applicability in autonomous driving.

\section{Conclusion and future work}
This paper presents a Perception, Language, and Action (PLA) framework for autonomous driving, integrating multi-sensor fusion (radar, LiDAR, camera) with a GPT-augmented VLA agent for explainable, adaptive, and safety-bounded decision-making in complex urban environments, such as construction zones. The PLA system unifies perception, language, and action planning, achieving robust speed and trajectory prediction.

Future work will focus on enhancing steering control precision, optimizing real-time performance, and expanding validation to diverse scenarios, including rare edge cases, to advance trustworthy and human-aligned autonomous systems. To bridge the gap between simulation and real-world testing, we plan to integrate our framework with AutoFrame \cite{kirchner2024autoframe}, a framework enabling low-cost, hardware-in-the-loop evaluation in real vehicles, which will refine steering accuracy and system reliability. Additionally, we aim to leverage LLM-based toolchains, such as those in \cite{petrovic2025genai}, to generate dynamic driving scenarios from freeform textual requirements. This approach will enable tailored testing beyond standard datasets, offering fine-grained control over environmental conditions, vehicle configurations, and system constraints, thus improving generalization and robustness.
%\clearpage
%\printglossaries
\section*{Acknowledgment}
This research was funded by the Federal Ministry of Research, Technology and Space of Germany as part of the CeCaS project, FKZ: 16ME0800K.
\printbibliography

\end{document}